\documentclass[a4paper]{SVProc}
\usepackage[utf8]{inputenc}

\usepackage[utf8]{inputenc} % allow utf-8 input
\usepackage[T1]{fontenc}    % use 8-bit T1 fonts
\usepackage{hyperref}       % hyperlinks
\usepackage{url}            % simple URL typesetting
\usepackage{booktabs}       % professional-quality tables
\usepackage{amsfonts}       % blackboard math symbols
\usepackage{nicefrac}       % compact symbols for 1/2, etc.
\usepackage{microtype}      % microtypography
\usepackage{bbm}
\usepackage{epsfig}
\usepackage{graphicx}
\usepackage{xcolor}

\title{Nonprehensile Riemannian Motion Predictive Control}
\author{Hamid Izadinia \and Byron Boots \and Steven M. Seitz}
\institute{University of Washington}%\\
\begin{document}

\maketitle

\begin{abstract}
    Nonprehensile manipulation involves long horizon underactuated object interactions and physical contact with different objects that can inherently introduce a high degree of uncertainty. In this work, we introduce a novel Real-to-Sim reward analysis technique, called Riemannian Motion Predictive Control (RMPC), to reliably imagine and predict the outcome of taking possible actions for a real robotic platform. Our proposed RMPC benefits from Riemannian motion policy and second order dynamic model to compute the acceleration command and control the robot at every location on the surface. Our approach creates a 3D object-level recomposed model of the real scene where we can simulate the effect of different trajectories. We produce a closed-loop controller to reactively push objects in a continuous action space. We evaluate the performance of our RMPC approach by conducting experiments on a real robot platform as well as simulation and compare against several baselines. We observe that RMPC is robust in cluttered as well as occluded environments and outperforms the baselines. 
\end{abstract}

\section{Introduction}

Nonprehensile object manipulation is an important and yet challenging robotic task that has been studied for more than three decades~\cite{mason1981}, yet remains an area of active research~\cite{li2018push,nair2017combining}. Nonprehensile manipulation involves long horizon underactuated object interactions and physical contact with different objects that can inherently introduce a high degree of uncertainty. Suppose we want to push an object on a  surface cluttered with many other objects with various shapes and properties. In such a scenario, predicting the outcome of a push action is not trivial as pushing an object not only affects its own status but also can change the status of other objects. 

Traditionally, motion planning  and search algorithms were adopted for nonprehensile manipulation. These algorithms traditionally have high computational cost and rely on full knowledge of a constrained workspace to produce a sequence of kinematic actions~\cite{mason1981}. To address the computational complexity brought by uncertainty, the quasi-static planners aim to predict the motion of pushed objects by assuming a physical model of the world involving shape, friction and objects center of mass~\cite{lynch1992manipulation}.

More recently, deep learning has been adopted for learning simple robotic manipulation skills using visual sensory input in a controlled scenario~\cite{levine2016end,agrawal2016learning,levine2018learning}. Central to their approach is to train a deep neural network model in a data-driven fashion to predict the cost of applying different robotic actions for accomplishing tasks. While it is essential to reliably predict the cost of taking possible actions in a real robotic task, it is challenging to obtain the data and the supervision needed for training such cost function in less constrained scenarios.

Instead of estimating the cost of different actions, can we instead simulate the real environment in an on-line closed-loop such that we can compute the outcome of taking possible actions inside the simulation? 
In this paper, we address this question by proposing to analyse and compute rewards through sensing and simulating the real world environment using 3D object-level geometric recomposition, a Riemannian Motion Policy (RMP), and second order dynamics. Our Real-to-Sim reward analysis technique, called Riemannian Motion Predictive Control (RMPC), simulates the effect of different trajectories and action outcomes on a real robotic platform and creates a closed-loop reactive controller to push objects in a continuous action space. To control the robot at every location on the surface, we incorporate a Riemannian Motion Policy~\cite{cheng2018rmpflow} and second order dynamic model to compute the acceleration command.

Our goal is to produce an efficient second order dynamic system controller that maintains a continuous trajectory while preserving correct kinematics and smooth dynamics for nonprehensile object manipulation, in the presence of a high amount of uncertainty. To accomplish this goal, we focus on leveraging simulation to predict parameters of our second order controller. Instead of estimating the accurate physical parameters of the scene such as center of mass and surface friction we apply a reactive controller that deploys closed-loop feedback. Consequently, we are capable of correcting the motion if, due to uncertainties involved, the object deviates from the foreseen trajectory.

\section{Related work}

Conventionally, planning algorithms has been deployed for solving nonprehensile manipulation by incorporating search techniques to create a trajectory of kinematic movements for a robot arm to push or rearrange objects~\cite{pushplan,stilman2005navigation,johnson2016convergent}. To push various objects on simple trajectories,  ~\cite{hermans2013learning} learns to predicts suitable contact points by optimizing a scoring function trained on histogram features. To reach objects in clutter,~\cite{killpack2016model} deployed Model Predictive Control (MPC)~\cite{wang2009fast} to model the dynamics of a robot arm in contact with the environment. Compared to these past methods, our approach is more general and more efficient and the task we consider is more challenging and involves multiple object interactions.

More recently, deep learning has been incorporated for learning simple manipulation skills with a robot arm. 
~\cite{agrawal2016learning} predict the outcome of a pushing action by training a network on pairs of RGB images of before and after push action. This work is extended in ~\cite{nair2017combining} to learn a forward models for manipulating a rope with a robot arm in a supervised fashion. Using a series of observations via camera images, ~\cite{li2018push} tracks a history of push interactions and learns a model for pushing in scenarios with only a single object on the table and quantizing the action space. In contrast, our method is designed to be applicable in complex scenarios involving several objects and our controller produces continues actions.

Our work is related to recent efforts for deploying 3D and physics simulation environments for efficiently learning robotic policy. Several past works deployed domain adaptation to transfer policies learned in simulation to the real world~\cite{rusu2017sim,bousmalis2018using}. Domain randomization was proposed in~\cite{sadeghi2017cadrl} to train highly generalizable robotic policies inside a randomized simulation environment and was shown to be successfully applicable in various robotic applications~\cite{james2017transferring,tremblay2018training,andrychowicz2020learning}. In~\cite{bousmalis2018using} pixel-level image generation was incorporated to create auxiliary source of data.~\cite{james2019sim} learned to translate real and simulation data into a canonical representation to be used as robot observation. Several past works incorporated real images as a complementary source of data for learning control policies inside simulation that can be transferred to the real world~\cite{bousmalis2018using,sadeghi2018sim2realservo,sadeghi2019divis}. In contrast to all these works, we generate the simulation environment by recomposing the 3D scene and incorporating simulation physics environment as a computational model for predicting the outcome of robot action trajectories in a closed loop. To our knowledge, closed loop 3D scene recomposition and simulated action look ahead has not been addressed before for efficient policy evaluation inside simulation that is applicable in the real world.

\section{Approach}
\label{sec:rmpc_methods}

\begin{figure}[t]
\centering
\includegraphics[width=.99\linewidth]{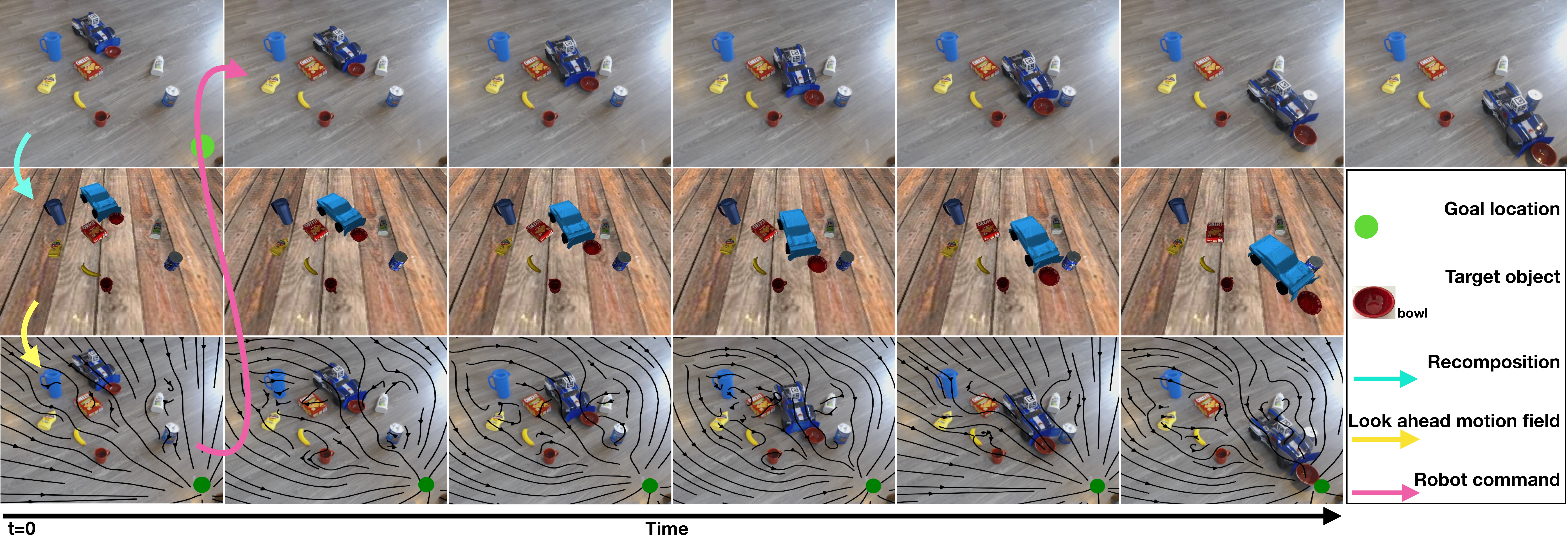}
\vspace{-0.05in}
\caption{\small
   \label{fig:rmpc_overview}
  Overview of our approach. {\bf Top row} shows the rgb snapshots of the point cloud stream input to our model which we use to recompose the scene into a simulated 3D scene ({\bf middle row}) and generate look ahead motion field ({\bf bottom row}).}
\vspace{-0.2in}
\end{figure}

We assume a nonprehensile robotic manipulation setup with $N$ objects $o_{i, i\in\{1,...,N\}}$. Each object has 5 degree-of-freedom (5-DoF) defined by its rotation $R\in \mathbb{SO}(3)$ and translation $T \in \mathbb{R}^2$ and can take any arbitrary pose $p$ in $\mathbb{SE}(2)$ on the plane. All the objects are movable on the plane and can be displaced from a location and pose to another physically feasible pose on the plane so at any time $t$, the state of the world $s_t$ encapsulates the pose of the objects in the scene $s_t=(p_1,...,p_N)$. 
Given a target object $o_c$ and a goal location $l_g=(x_g,y_g)$ as inputs, we want to control the target object $o_c$ by pushing it to $l_g$. To accomplish that, we need to find the best sequence $\tau$~of robot actions $\tau=\{u_0,u_1,...,u_t\}$ that pushes $o_c$ to $l_g$ while satisfying the constraints such as avoiding collision with other objects as well as maximizing a reward function. Each robot action $u$ is defined by the velocity and acceleration of a push by a robot end-effector, i.e. $u_t = (v_x,v_y,a_x,a_y)$. At every iteration, our goal is to estimate a short trajectory of future actions $\tau$ that maximizes the expected sum of future rewards. By applying each action $u_t$ at state $s_t$, a reward $r_t(s_t,u_t)$ will be obtained which reflects how much the object has progressed in becoming closer to the goal location. If the action $u_t$ results in a collision between the controllable object $o_c$ and any of the objects in the scene $o_{k, k\in\{1,...,N\}, k\neq c}$ then a penalty would be considered. More formally, the reward at each step $r_t$ is defined as:
\vspace{-.1in}
\begin{equation}
\label{eq:reward}
   r_t(s_t,u_t) = \Delta(d_{o_c,l_g}^{t},d_{o_c,l_g}^{t-1})-  \sum_{i=1, i\neq p}^{N} d(p^{t}_i,p^{t-1}_i) 
\end{equation}
\vspace{-.1in}

\noindent The first term in Equation~\ref{eq:reward} reflects how much the distance of $o_c$ to the goal location $d_{o_c,l_g}$ has changed between time $t-1$ and $t$ and $d$ denotes the Euclidean distance. The second term of Equation~\ref{eq:reward} quantifies the collision event that might occur by applying action $u_t$. 
During one step of push action, if the controllable object $o_c$ collides with any other object $o_i$, it will result in an unwanted change of 5-DoF pose in $o_i$. To quantify the collision event, we compare the pose of each object $p_i$ between the time $t$ and $t-1$ by computing the Euclidean distance $d$ between $p^{t}_i$ and $p^{t-1}_i$. 
Accordingly, the reward of a trajectory $\tau$ with fixed horizon   $H$ is computed as: $R_{\tau}=\sum_{j=t}^{t+H} \gamma^{j-t} r_t$ with the decay factor of $\gamma$.

\begin{figure}[t]
\centering
\includegraphics[width=.99\linewidth]{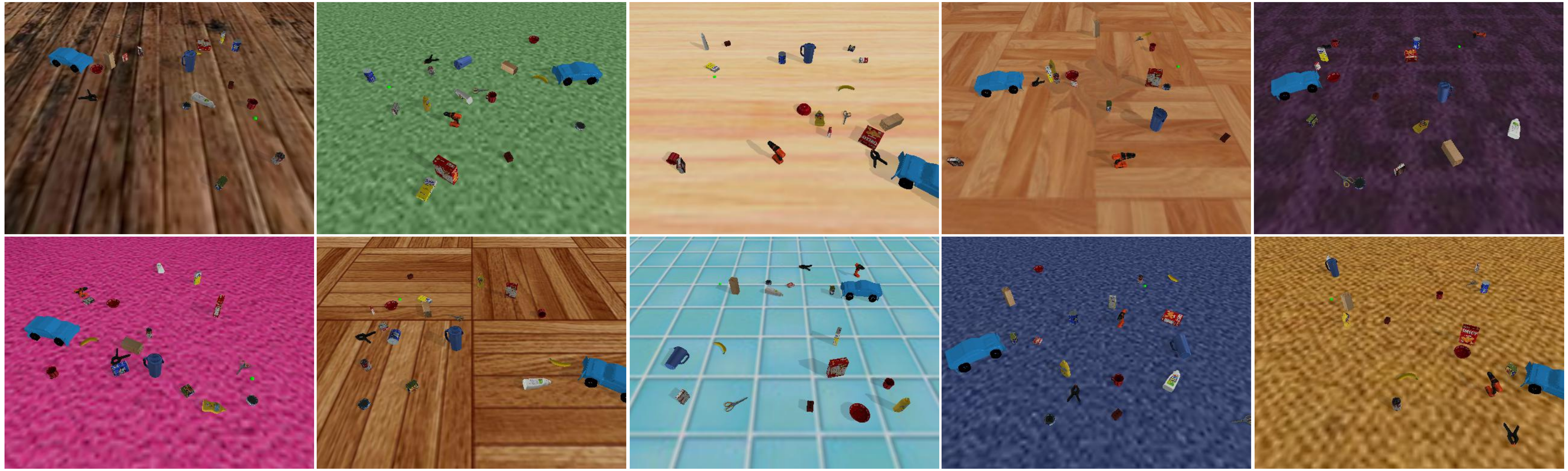}
\vspace{-0.05in}
\caption{\small
   \label{fig:sample_task}
  Sample pushing tasks for RC-car.}
\vspace{-0.3in}
\end{figure}

\subsection{Closed-Loop Control}
We use RMPflow~\cite{cheng2018rmpflow} to compute the sequence of robot actions based on the current robot position and velocity and the configuration of scene objects. The RMPflow combines local policies in their coordinate system and computes a global policy for the robot motion. We use attractor policy for pushing to the goal and collision avoidance policy for avoiding obstacles. By combining these local motion policies the potential field of robot action is computed in continuous space which determines the robot global motion policy at every location.
Suppose the average dimension $o_c$ along its width and length is $\bar{m}_{o_c}$ and its location on the plane is $l_{o_c}=(x_{o_c},y_{o_c})$. The local policy in the coordinate of the target object rotated by the orientation of the RMP global policy for an immediate pushing action is defined by $(v_x, v_y) = (\alpha_1 x^2 - \alpha_2 y^2 - \alpha_3 ( \frac{\bar{m}_{o_c}}{2})^2, \alpha_4xy)$ and $\alpha_i$s are the hyperparameters of our local potential field.

\subsection{Riemannian Motion Predictive Control}
Using the computed RMPflow~\cite{cheng2018rmpflow} we can select an action $u_t$ at each time step $t$ to find the optimum trajectory to push $o_c$ to $l_g$. However, in dynamic systems with a high degree of uncertainty selecting the actions purely based on RMPflow may not be optimal specially in underactuated tasks~\cite{lynch1996dynamic} such as ours where the target object makes and breaks contact the with robot end-effector and other objects. 
Therefore, we propose Riemannian Motion Predictive Control (RMPC) where instead of fixing a single parameter for computing the RMPflow, we consider a range of weights for each node. By sampling from the different node weights, we produce different pushing trajectories at every time step $t$ and select the next best action based on the trajectory with the highest estimated reward using Equation~\ref{eq:reward}.

\begin{figure}[t]
\centering
\includegraphics[width=.99\linewidth]{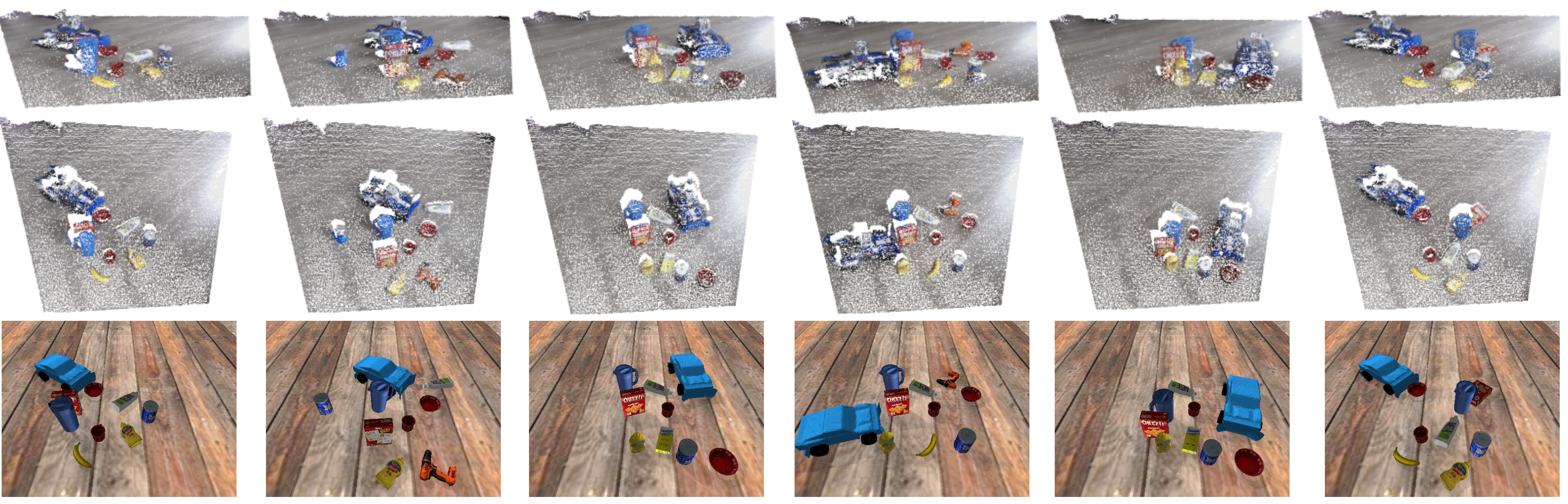}
\vspace{-0.05in}
\caption{\small
   \label{fig:recompose_visualization}
 Examples of fully automatic object-level scene recomposition from the point cloud. The {\bf first} and {\bf second} rows show the side view and the top view of the captured point cloud. The {\bf third} row shows our object-level scene recomposition.}
\vspace{-0.25in}
\end{figure}

Designing an accurate reward function and learning to predict  the outcome of different action sequences is challenging due to inefficiencies in collecting real robot data and inherent uncertainties in under-actuated and non-prehensile object manipulation. To alleviate those challenges, our proposed method estimates the reward of different sampled action trajectories by recomposing the 3D scene and imagining the outcome of different action trajectories inside a 3D physics simulation. 
We leverage the power of 3D physics simulation to look ahead the outcome of pushing an object along different trajectories by proposing to recompose the 3D scene at each step during the trajectory. In our setup, we have a depth camera that captures the scene and we incorporate sensory point cloud input for object-level scene recomposition. Using the captured point cloud, we detect 3D objects in the scene.  Once we recomposed the scene using object-level 3D shapes, we migrate the scene to a physics simulator to evaluate the outcome of different pushing trajectories. 
To train a 3D object detector that is robust to clutter and high amount of occlusion, we generate a high volume of diverse synthetic data in a simulation environment inspired from prior work of~\cite{izadinia2020scene} and domain randomization~\cite{sadeghi2017cadrl}. We create a huge number of diverse scenes by randomly dropping objects in a physic simulation. Such data generation approach will capture the realistic physical arrangement of objects on a surface and results in learning a more efficient 3D object detector.  Figure~\ref{fig:rmpc_overview} illustrates the overview of our approach where scenes captured by point clouds are recomposed in a 3D simulation environment where we can compute the look ahead motion field. 

\subsection{Implementation Details} 

We implemented our method using PyTorch~\cite{NEURIPS2019_9015} and used Bullet physics engine for simulating the scenes~\cite{coumans2015bullet}. We generate over 1 million simulated scenes diversified with random object arrangements and floor texturing to train our 3D object detector and Figure~\ref{fig:sample_task} provides several examples of the generated scenes. The speed of our recomposition step is 15 frames per second. We set the hyperparameter of our local field to $\alpha_i$s$=1$ in all our experiments.

\section{Experiments}

We conduct experiments both in simulation and in a real world to
evaluate the performance of our proposed approach for nonprehensile manipulation. To evaluate the performance of our proposed approach for nonprehensile manipulation we conduct experiments both in simulation and in a real robot platform.

\begin{figure}[t]
\centering
\includegraphics[width=.99\linewidth]{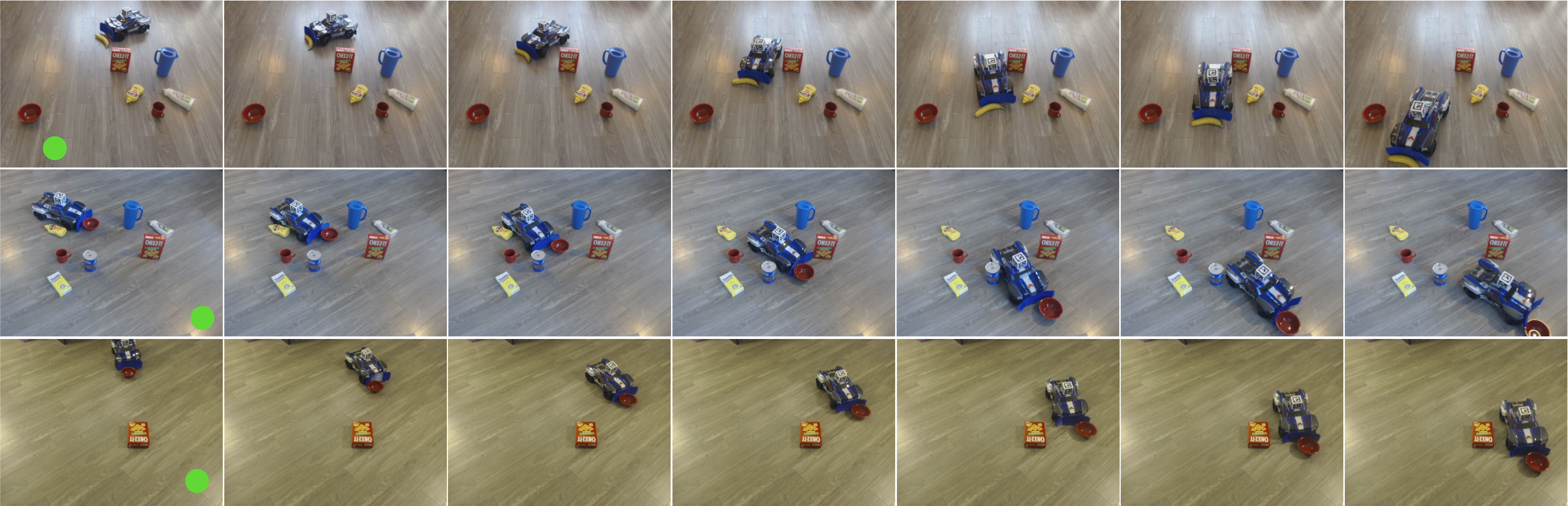}
\vspace{-0.05in}
\caption{\small
   \label{fig:real_exp_traj}
 Real world RC-car robot action trajectory generated by proposed RMPC. Each row shows a trial with a different initial location of the target object, other objects as obstacles. The goal location is indicated with a green dot.}
\vspace{-0.17in}
\end{figure}

\subsection{Experimental Setup}
In our evaluations, We use objects of the YCB dataset~\cite{calli2015benchmarking} in simulation and in the real world experiments. For real robot experiments, we build a real RC-car hardware platform on a 1/10 chassis featuring a 4$\times$4 suspension, and non-flat tires inspiring from the prototypes introduced in~\cite{mitcar,srinivasa2019mushr}. To control the RC-car we use move commands that define the velocity and steering direction of the RC-car through a ROS interface. In the real experiments, we localize the RC-car robot using the April tag~\cite{apriltag} while we detect and localize other scene objects by 3D object detection as explained in Section~\ref{sec:rmpc_methods}.

\begin{figure}[t]
\centering
\includegraphics[width=.99\linewidth]{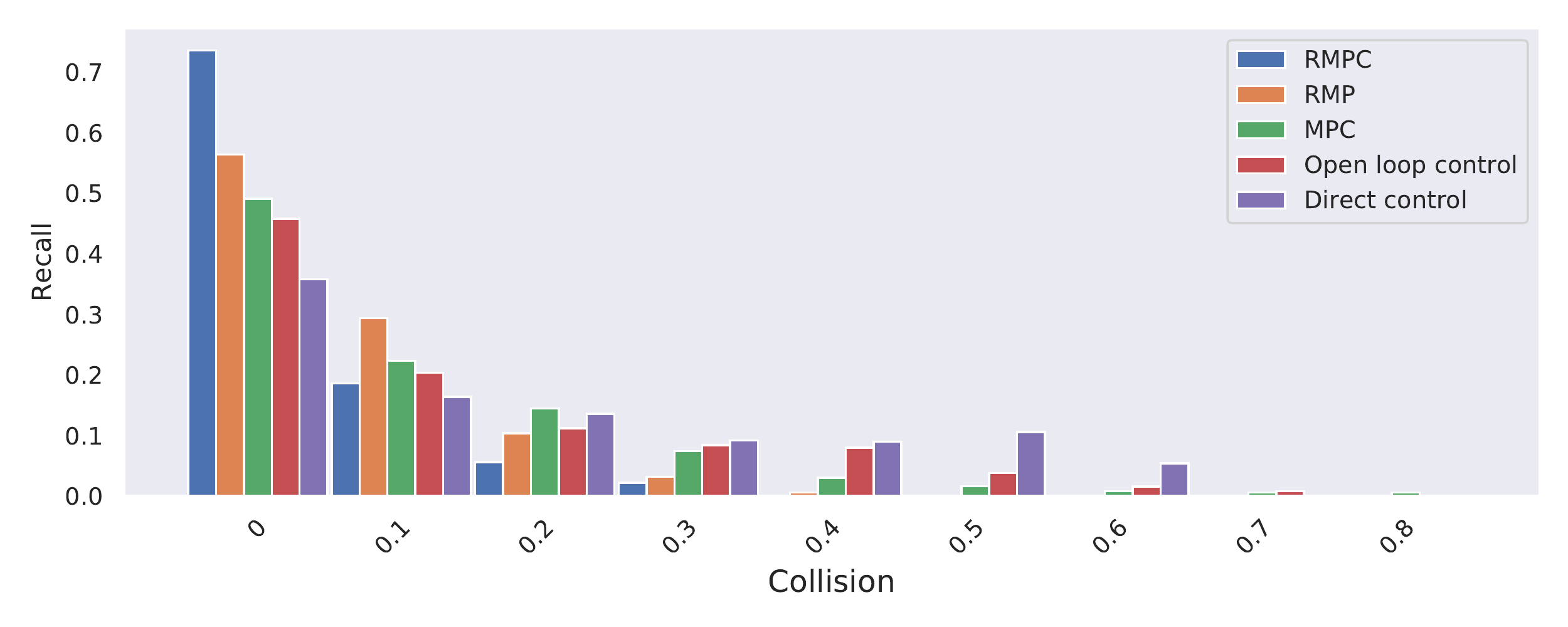}
\vspace{-0.15in}
\caption{\small
   \label{fig:collision}
  Results of collision rate on held out test set using proposed Riemannian Motion Predictive Control (RMPC) on RC-car robot platform for pushing target object to goal location compare to prior works. (higher recall in lower collision rate is better).}
\vspace{-0.1in}
\end{figure}

\subsection{Results on Simulation}
\label{sec:sim_exp}
To evaluate our proposed motion policy we generate a held-out test set of scenes with a different combination of objects and various arrangements in simulation. To produce diverse scenarios, we randomly place objects in random locations of the workspace surface area and randomly select the target object and goal location in simulation. Figure~\ref{fig:sample_task} shows several examples of push tasks at test time. We use 500 held-out test scenes on our simulated RC-car robot platform and compare performance based on the collision rate. We compute the recall rate at the $k\%$-collision ratio as evaluation criteria. For each trial, the collision ratio is computed by dividing the number of collision events by the total length of the trajectory. A higher recall rate at a lower collision ratio indicates better performance.

We compare the performance of our RMPC method against several baselines:

{\bf Riemannian Motion Predictive Control (RMPC):} Our proposed RMPC recomposes scene from input point cloud and computes different trajectories using RMPflow with different weights. The next best action is selected based on future reward obtained from Equation\ref{eq:reward} 

{\bf Riemannian Motion Policy (RMP):} RMP generates the action trajectory based on the computed RMPflow with a fixed parameter similar to~\cite{cheng2018rmpflow}. 

{\bf Model Predictive Control (MPC):} 
MPC samples different trajectories at each step and selects the next action from the trajectory with the smallest cost in a closed-loop. 
The cost function is computed based on the distance of the target object and the location of the obstacles so  the trajectory that moves closer to the obstacles is of higher cost. 

{\bf Open-loop control:} This baseline is similar to MPC except that it computes the whole trajectory from the initial location of the target object to the goal location and applies action sequence in an open-loop.

{\bf Direct control:} This baseline moves the target object toward goal location without considering collision avoidance and recomputes the action in closed-loop.

\begin{figure}[t]
\centering
\includegraphics[width=.99\linewidth]{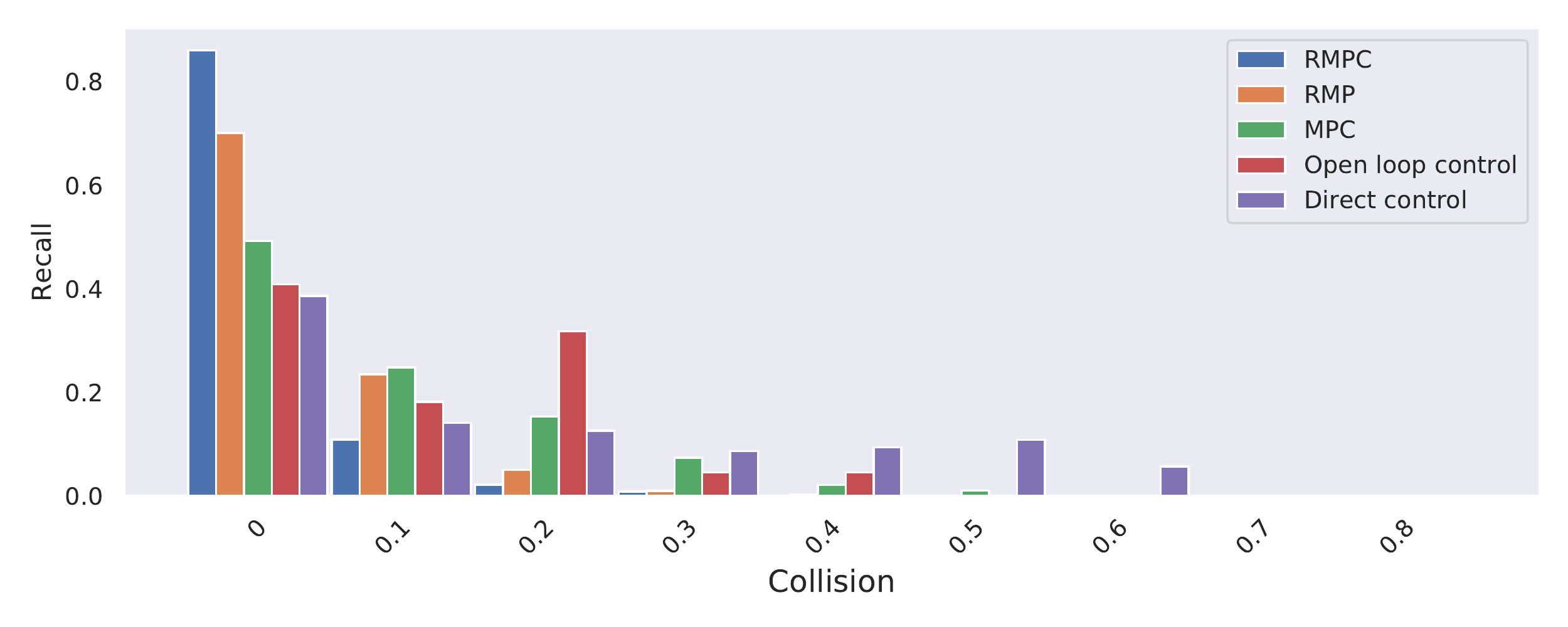}
\vspace{-0.15in}
\caption{\small
   \label{fig:collision_success}
  Collision rate of proposed RMPC controller on RC-car robot platform for successful trajectories where target object is pushed close to the goal location on held out test set compared to prior works. (higher recall in lower collision rate is better).}
\vspace{-0.1in}
\end{figure}

Comparison results are summarized in Figures~\ref{fig:collision} and~\ref{fig:collision_success} and show that our proposed RMPC method outperforms all of the baselines and experiences less number of collisions during the trails. Hence RMPC obtains higher recall at lower collision ratios.

\subsection{Results on Real-World}

In all real experiments we capture the scene point cloud using a depth camera. Given the point cloud of a scene, our 3D object detection produces the 3D bounding boxes around each object with their 3D poses as well as category labels from which we can recompose the scene in the physics simulation. Figure~\ref{fig:recompose_visualization} depicts several examples of the recomposed real scene in the 3D simulator. As can be seen, our scene recomposition produces simulated scenes that are identical to the real world in terms of object scene arrangement, various object categories, and their poses. Also, our scene recomposition is robust to a high amount of clutter and occlusion in the scene. 

We quantitatively compare our proposed RMPC with RMP which we observed to be the strongest baseline in Section~\ref{sec:sim_exp}. We evaluate the performance by computing the percentage of non-collision push actions in different trials. We evaluate the performance in $32$ different real scene arrangements and obtained the average collision rate of $17.9\%$ for RMPC and $35.4\%$ average collision for RMP. Figure~\ref{fig:real_exp_traj} shows the example trajectories of real RC-car pushing tasks in three different scenarios. As Figure~\ref{fig:real_exp_traj} shows, our proposed RMPC method produces a closed-loop sequence to successful actions and pushes the controllable object to the goal location(shown by green dot) without colliding to with obstacles in the scene where pushing to a goal location could be accomplished without a collision event.

\section{Conclusion}

Solving for the parameters of a second order dynamic system controller is challenging when the task involves objects of various properties and complex manipulation under high uncertainty specially in nonprehensile settings. In this paper, we propose a Real-to-Sim approach for analysing the reward of action trajectory in simulation and estimating the parameters of a second order dynamic system controller for nonprehensile manipulation task. With our controller, we produce a closed-loop reactive motion policy. We consider the complex task of pushing objects on a surface in a complex environment with high uncertainty and demonstrate the efficiency of our controller by comparing it to the prior works on a set of nonprehensile manipulation tasks in simulation and real-world RC-car robot platform.

\vspace{-0.05in}
\section*{Acknowledgments}
\vspace{-0.05in}

This work was supported in part by the University of Washington Animation Research Labs and Google.

\bibliographystyle{splncs03}
\bibliography{Hamid_RMPC}

\begin{thebibliography}{10}
\providecommand{\url}[1]{\texttt{#1}}
\providecommand{\urlprefix}{URL }

\bibitem{mitcar}
The MIT RACECAR (2016), \url{https://mit-racecar.github.io}

\bibitem{agrawal2016learning}
Agrawal, P., Nair, A.V., Abbeel, P., Malik, J., Levine, S.: Learning to poke by
  poking: Experiential learning of intuitive physics. In: NeurIPS (2016)

\bibitem{andrychowicz2020learning}
Andrychowicz, O.M., Baker, B., Chociej, M., Jozefowicz, R., McGrew, B.,
  Pachocki, J., Petron, A., Plappert, M., Powell, G., Ray, A., et~al.: Learning
  dexterous in-hand manipulation. The International Journal of Robotics
  Research  (2020)

\bibitem{bousmalis2018using}
Bousmalis, K., Irpan, A., Wohlhart, P., Bai, Y., Kelcey, M., Kalakrishnan, M.,
  Downs, L., Ibarz, J., Pastor, P., Konolige, K., et~al.: Using simulation and
  domain adaptation to improve efficiency of deep robotic grasping. In: ICRA
  (2018)

\bibitem{calli2015benchmarking}
Calli, B., Walsman, A., Singh, A., Srinivasa, S., Abbeel, P., Dollar, A.M.:
  Benchmarking in manipulation research: The ycb object and model set and
  benchmarking protocols. arXiv:1502.03143  (2015)

\bibitem{cheng2018rmpflow}
Cheng, C.A., Mukadam, M., Issac, J., Birchfield, S., Fox, D., Boots, B.,
  Ratliff, N.: Rmpflow: A computational graph for automatic motion policy
  generation. In: International Workshop on the Algorithmic Foundations of
  Robotics (2018)

\bibitem{pushplan}
{Cosgun}, A., {Hermans}, T., {Emeli}, V., {Stilman}, M.: Push planning for
  object placement on cluttered table surfaces. In: IEEE/RSJ International
  Conference on Intelligent Robots and Systems (2011)

\bibitem{coumans2015bullet}
Coumans, E.: Bullet physics simulation. In: ACM SIGGRAPH 2015 Courses (2015)

\bibitem{hermans2013learning}
Hermans, T., Li, F., Rehg, J.M., Bobick, A.F.: Learning contact locations for
  pushing and orienting unknown objects. In: IEEE-RAS international conference
  on humanoid robots (humanoids) (2013)

\bibitem{izadinia2020scene}
Izadinia, H., Seitz, S.M.: Scene recomposition by learning-based icp. In: CVPR
  (2020)

\bibitem{james2017transferring}
James, S., Davison, A.J., Johns, E.: Transferring end-to-end visuomotor control
  from simulation to real world for a multi-stage task. arXiv:1707.02267
  (2017)

\bibitem{james2019sim}
James, S., Wohlhart, P., Kalakrishnan, M., Kalashnikov, D., Irpan, A., Ibarz,
  J., Levine, S., Hadsell, R., Bousmalis, K.: Sim-to-real via sim-to-sim:
  Data-efficient robotic grasping via randomized-to-canonical adaptation
  networks. In: CVPR (2019)

\bibitem{johnson2016convergent}
Johnson, A.M., King, J.E., Srinivasa, S.: Convergent planning. IEEE Robotics
  and Automation Letters  (2016)

\bibitem{killpack2016model}
Killpack, M.D., Kapusta, A., Kemp, C.C.: Model predictive control for fast
  reaching in clutter. Autonomous Robots  (2016)

\bibitem{levine2016end}
Levine, S., Finn, C., Darrell, T., Abbeel, P.: End-to-end training of deep
  visuomotor policies. The Journal of Machine Learning Research  (2016)

\bibitem{levine2018learning}
Levine, S., Pastor, P., Krizhevsky, A., Ibarz, J., Quillen, D.: Learning
  hand-eye coordination for robotic grasping with deep learning and large-scale
  data collection. The International Journal of Robotics Research  (2018)

\bibitem{li2018push}
Li, J.K., Lee, W.S., Hsu, D.: Push-net: Deep planar pushing for objects with
  unknown physical properties. In: Robotics: Science and Systems (2018)

\bibitem{lynch1992manipulation}
Lynch, K.M., Maekawa, H., Tanie, K.: Manipulation and active sensing by pushing
  using tactile feedback. In: IROS (1992)

\bibitem{lynch1996dynamic}
`Lynch, K.M., Mason, M.T.: Dynamic underactuated nonprehensile manipulation.
  In: IROS (1996)

\bibitem{mason1981}
{Mason}, M.T.: Compliance and force control for computer controlled
  manipulators. IEEE Transactions on Systems, Man, and Cybernetics  (1981)

\bibitem{nair2017combining}
Nair, A., Chen, D., Agrawal, P., Isola, P., Abbeel, P., Malik, J., Levine, S.:
  Combining self-supervised learning and imitation for vision-based rope
  manipulation. In: ICRA (2017)

\bibitem{apriltag}
Olson, E.: {AprilTag}: A robust and flexible visual fiducial system. In: ICRA
  (2011)

\bibitem{NEURIPS2019_9015}
Paszke, A., Gross, S., Massa, F., Lerer, A., Bradbury, J., et~al.: Pytorch: An
  imperative style, high-performance deep learning library. In: NeurIPS (2019)

\bibitem{rusu2017sim}
Rusu, A.A., Ve{\v{c}}er{\'\i}k, M., Roth{\"o}rl, T., Heess, N., Pascanu, R.,
  Hadsell, R.: Sim-to-real robot learning from pixels with progressive nets.
  In: CoRL (2017)

\bibitem{sadeghi2019divis}
Sadeghi, F.: {DIViS}: Domain invariant visual servoing for collision-free goal
  reaching. In: RSS (2019)

\bibitem{sadeghi2017cadrl}
Sadeghi, F., Levine, S.: {CAD2RL}: Real single-image flight without a single
  real image. In: RSS (2017)

\bibitem{sadeghi2018sim2realservo}
Sadeghi, F., Toshev, A., Jang, E., Levine, S.: Sim2real viewpoint invariant
  visual servoing by recurrent control. In: CVPR (2018)

\bibitem{srinivasa2019mushr}
Srinivasa, S.S., Lancaster, P., Michalove, J., Schmittle, M., Rockett, C.S.M.,
  Smith, J.R., Choudhury, S., Mavrogiannis, C., Sadeghi, F.: Mushr: A low-cost,
  open-source robotic racecar for education and research. arXiv preprint
  arXiv:1908.08031  (2019)

\bibitem{stilman2005navigation}
Stilman, M., Kuffner, J.J.: Navigation among movable obstacles: Real-time
  reasoning in complex environments. International Journal of Humanoid Robotics
   (2005)

\bibitem{tremblay2018training}
Tremblay, J., Prakash, A., Acuna, D., Brophy, M., Jampani, V., Anil, C., To,
  T., Cameracci, E., Boochoon, S., Birchfield, S.: Training deep networks with
  synthetic data: Bridging the reality gap by domain randomization. In: CVPRW
  (2018)

\bibitem{wang2009fast}
Wang, Y., Boyd, S.: Fast model predictive control using online optimization.
  IEEE Transactions on control systems technology  (2009)

\end{thebibliography}
\end{document}